\documentclass[10pt,twocolumn,letterpaper]{article}

\usepackage{cvpr}              % To produce the CAMERA-READY version
\definecolor{cvprblue}{rgb}{0.21,0.49,0.74}
\usepackage[pagebackref,breaklinks,colorlinks,allcolors=cvprblue]{hyperref}
\usepackage[accsupp]{axessibility}
\usepackage{booktabs}       % Professional-quality tables
\usepackage{amsfonts}       % Blackboard math symbols
\usepackage{nicefrac}       % Compact symbols for 1/2, etc.
\usepackage{microtype}      % Microtypography
\usepackage{xcolor}         % Colors
\usepackage{graphicx}       % Provides \resizebox
\usepackage{amsmath}
\usepackage{amssymb}
\usepackage{enumitem}
\usepackage{colortbl}       % Cell background colors
\usepackage{pifont}         % Provides \cmark and \xmark
\usepackage{multirow}
\usepackage{caption}
\usepackage{multibib}
\newcites{Supp}{Supplementary References}

\newcommand{\cmark}{\ding{51}} % Checkmark
\newcommand{\xmark}{\ding{55}} % Cross
\definecolor{lakegreen}{HTML}{CCFAEB}
\definecolor{lakeblue}{HTML}{0077FF}

\def\confName{CVPR}
\def\confYear{2025}

\title{MaRI: Material Retrieval Integration across Domains}

\author{
    Jianhui Wang\textsuperscript{1*} \hspace{0.25em}
    Zhifei Yang\textsuperscript{2*} \hspace{0.25em}
    Yangfan He\textsuperscript{3*} \hspace{0.25em}
    Huixiong Zhang\textsuperscript{1\dag} \hspace{0.25em}
    Yuxuan Chen\textsuperscript{4} \hspace{0.25em}
    Jingwei Huang\textsuperscript{5\dag} \\
    \textsuperscript{1}University of Electronic Science and Technology of China \quad 
    \textsuperscript{2}Peking University \quad \\
    \textsuperscript{3}University of Minnesota-Twin Cities \quad 
    \textsuperscript{4}Fudan University \quad 
    \textsuperscript{5}Tencent Hunyuan3D \\
    { \textbf{\url{https://jianhuiwemi.github.io/MaRI}}} \\
}

\begin{document}

\twocolumn[{%
    \maketitle
    \vspace{-20pt}
    \begin{center}
      \includegraphics[width=1.0\linewidth]{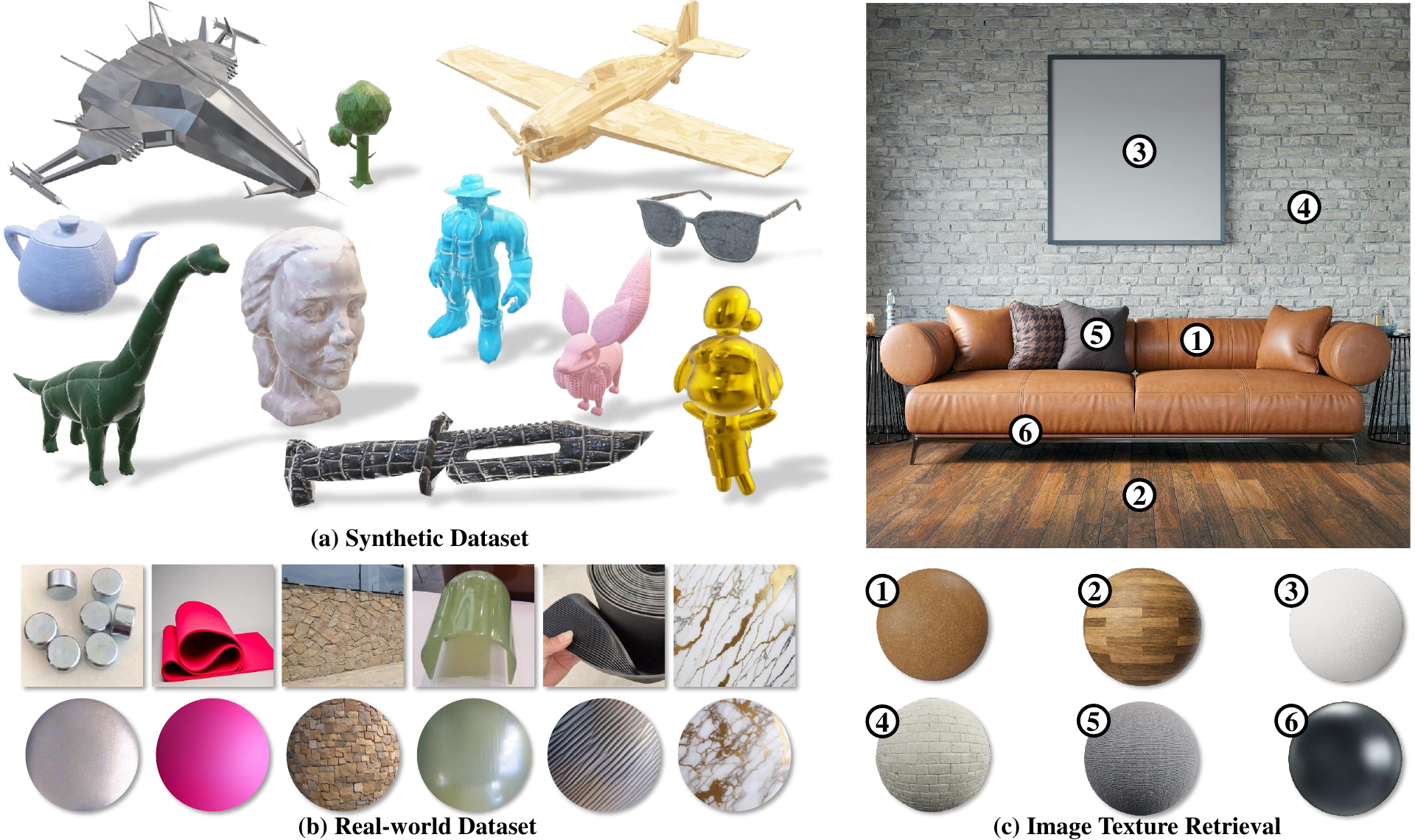}
      \captionof{figure}{
        Examples from the MaRI Gallery, showcasing (a) synthetic and (b) real-world datasets we constructed.
        (c) MaRI: A groundbreaking framework for accurately retrieving textures from images, bridging the gap between visual representations and material properties.
      }
      \label{fig:head}
    \end{center}
}]

\def\thefootnote{*}\footnotetext{Equal contribution.}\def\thefootnote{\arabic{footnote}}
\def\thefootnote{\dag}\footnotetext{Corresponding author.}\def\thefootnote{\arabic{footnote}}

\begin{abstract}
Accurate material retrieval is critical for creating realistic 3D assets. Existing methods rely on datasets that capture shape-invariant and lighting-varied representations of materials, which are scarce and face challenges due to limited diversity and inadequate real-world generalization. Most current approaches adopt traditional image search techniques. They fall short in capturing the unique properties of material spaces, leading to suboptimal performance in retrieval tasks. Addressing these challenges, we introduce MaRI, a framework designed to bridge the feature space gap between synthetic and real-world materials. MaRI constructs a shared embedding space that harmonizes visual and material attributes through a contrastive learning strategy by jointly training an image and a material encoder, bringing similar materials and images closer while separating dissimilar pairs within the feature space. To support this, we construct a comprehensive dataset comprising high-quality synthetic materials rendered with controlled shape variations and diverse lighting conditions, along with real-world materials processed and standardized using material transfer techniques. Extensive experiments demonstrate the superior performance, accuracy, and generalization capabilities of MaRI across diverse and complex material retrieval tasks, outperforming existing methods.
\end{abstract}
\vspace{-20pt}    
\section{Introduction}
\label{sec:introduction}
The creation of realistic appearances is a crucial aspect of 3D asset generation, with accurate material reconstruction, particularly through high-quality materials in UV texture space, being key to achieving photorealism \cite{guerrero2022matformer,zhou2022tilegen,hu2023generating,zhou2023photomat,vecchio2024controlmat,vecchio2024matfuse}. This is especially important in applications like augmented reality (AR), virtual reality (VR), digital content creation, and industrial design, where the seamless integration of virtual objects into real-world environments depends on faithfully reproducing material properties \cite{li2018learning, deschaintre2021deeppolarizationimaging3d, sharma2023materialisticselectingsimilarmaterials, raistrick2023infinitephotorealisticworldsusing}. A fundamental challenge in these domains is aligning the information from the material space with that from the image space. The goal is to project both into a shared embedding space, enabling accurate comparison and retrieval of materials. Achieving this alignment is critical, as it allows for high-quality material searches that can accurately match visual inputs with corresponding material representations, leading to more realistic and context-aware renderings.

Material retrieval can theoretically be viewed as an image search problem, which suggests that the task should be relatively straightforward given the vast array of image search techniques available today, including vision transformers \cite{dosovitskiy2021imageworth16x16words}, DINOv2 \cite{oquab2024dinov2learningrobustvisual} and multimodal approaches like CLIP \cite{radford2021learningtransferablevisualmodels} and GPT-4V \cite{openai2024gpt4technicalreport}.
While the image space has been extensively explored and understood, applying image-based methods directly to material search often falls short. For example, some recent efforts have attempted to adapt image search techniques for material retrieval but the results have been suboptimal \cite{zhang2024mapatextdrivenphotorealisticmaterial, fang2024makeitrealunleashinglargemultimodal}.
We argue that the problem is caused by inherent differences between the material and image. As a result, material search and image search differ fundamentally since it requires capture a feature space specifically for materials properties including texture, reflectance, and surface roughness.
Unfortunately, such a well-defined feature space for materials is unavailable due to the lack of comprehensive datasets and effective material descriptors. The absence of a meaningful material embedding makes it challenging to achieve accurate retrieval, highlighting the need for learning a shared space that align material and image information for more effective searches.

To handle these challenges, we introduce MaRI—a novel framework inspired by CLIP that learns a joint embedding space for both materials and images. MaRI employs dual encoders, jointly trained to align material properties with visual features in a shared space, facilitating direct and efficient comparisons. By leveraging pre-trained DINOv2 models as the backbone for both the material and image encoders, MaRI preserves generalizability while fine-tuning only the final Transformer block to capture domain-specific nuances effectively.
We construct a large-scale dataset pairing images with materials, designed to capture both diversity and realism, for training a joint embedding.
Since such a dataset is unavailable, we adopt both synthesis and generative approach to automatically construct the dataset. During synthesis, we construct each data pair by sampling a material from a material gallery, associating it to an object from an object dataset, and render it with Blender to obtain an image that pairs with the material.
Although synthetic data alone yields satisfactory results for training, it still partially falls short in bridging the domain gap, as it cannot fully represent the diverse and nuanced appearances of real-world materials. To complement this, we introduce a generative approach that incorporate large-scale, unlabeled real-world image data and construct paired material with a material transfer technique (ZeST) \cite{cheng2024zestzeroshotmaterialtransfer}. As a result, the generative approach captures diverse material appearances under varying conditions through real images. It enables us to adopt a self-supervised learning framework, helping the model to learn robust material representations without being dependent on annotated datasets.

% \commentjw{possibly write another section introduce experiments}
The effectiveness of MaRI is validated through a series of experiments. These evaluations focus on two distinct datasets: one emphasizing retrieval within a gallery of synthetic materials the model was trained on, and another assessing generalization to unseen materials. The results demonstrate MaRI’s ability to bridge the domain gap between synthetic and real-world data, achieving significant improvements in both instance-level and class-level retrieval tasks. Our main contributions are summarized as:
\begin{itemize}
    \item We propose MaRI, a framework that learns a joint embedding space for visual and material properties, providing a new approach to material retrieval by aligning visual features with material characteristics.
    \item We construct a diverse dataset that spans various material types and conditions, supporting both synthetic and real-world material retrieval.
    \item We show the ability of MaRI’s retrieval pipeline to accurately scale across a wider variety of materials, achieving precise retrieval for complex and diverse material types.
\end{itemize}

\section{Related Work}
\textbf{Datasets for Material Understanding.} Over the years, various datasets have played a crucial role in advancing material understanding. Early works laid the foundation by capturing real-world material reflectance properties, providing basic 3D models, generating synthetic photorealistic images with ground truth annotations, and conducting similarity assessments for 3D reconstruction and material synthesis \cite{bell2013opensurfaces, chang2015shapenetinformationrich3dmodel, bell2015material, Ley2016SyB3RAR, Park_2018}. For instance, Flickr Material Database \cite{Liu-CVPR-10} contributed to material recognition by providing labeled images of real-world materials for classification tasks. Datasets like the Amazon-Berkeley Objects (ABO) \cite{collins2022abodatasetbenchmarksrealworld} with high-resolution 3D models and PBR materials, and MatSynth \cite{vecchio2024matsynthmodernpbrmaterials} and OmniObject3D \cite{wu2023omniobject3dlargevocabulary3dobject} with thousands of PBR materials and real-scanned objects, have significantly enhanced material diversity and realism.

Despite these advancements, challenges remain in achieving sufficient material diversity and integrating real-world and synthetic data. Existing datasets and methods have inspired us to construct a more diverse dataset, incorporating richer environmental contexts, complex shapes, and more detailed material information. AmbientCG \cite{ambientcg2023} provides a rich library of high-quality, PBR materials, designed for use in physically-based rendering workflows, while Objaverse \cite{deitke2022objaverseuniverseannotated3d} offers an extensive collection of 3D models for material application and visual tasks across synthetic environments. Additionally, the ZeST \cite{cheng2024zestzeroshotmaterialtransfer} method’s material transfer approach informs our efforts to capture diverse appearances under varying conditions.\\
\textbf{Material Generation and Retrieval.} Advances in material generation and retrieval have enhanced the realism of 3D content creation. Techniques like ControlMat \cite{vecchio2024controlmat} use diffusion models to generate high-resolution material maps, allowing precise control over surface properties such as roughness and normal maps. MatFuse \cite{vecchio2024matfuse} enables users to guide the material generation process through sketches, color palettes, or text prompts, providing greater flexibility in design. Similarly, Fantasia3D \cite{chen2023fantasia3ddisentanglinggeometryappearance} employs pixel-level optimization for detailed material representations, though it faces challenges in maintaining stability with complex geometries. Make-it-3D \cite{tang2023makeit3dhighfidelity3dcreation} uses a two-stage diffusion process to generate high-fidelity 3D content from single images, emphasizing texture refinement for realistic outputs. Tools like Material Palette \cite{lopes2023materialpaletteextractionmaterials} and Matlas \cite{ceylan2024matatlastextdrivenconsistentgeometry} further advance material manipulation by offering more control over varied environmental conditions and geometries.

Several studies \cite{10.1145/3306346.3323036, 10.1145/3450626.3459813} have focused on perceptual similarity. However, none directly address the challenge of retrieving accurate materials from real-world images. MaPa \cite{zhang2024mapatextdrivenphotorealisticmaterial} and Make-it-Real \cite{fang2024makeitrealunleashinglargemultimodal} integrate material retrieval within broader 3D generation workflows. MaPa utilizes GPT-4V \cite{openai2024gpt4technicalreport} for initial material categorization and CLIP \cite{radford2021learningtransferablevisualmodels} for refining the material assignment process, retrieving material graphs from predefined libraries, though it often lacks precision in handling fine textures and complex surfaces. Make-it-Real leverages GPT-4V to assign materials to segmented 3D models using SVBRDF \cite{haindl2013spatially} mappings, but its reliance on pre-annotated datasets limits adaptability to novel materials and environments. These limitations underscore the need for more robust frameworks. Our work addresses this by bridging the gap between real-world and synthetic data, capturing both fine textures and complex material details.
\begin{figure*}[t]
    \centering
    \includegraphics[width=\textwidth]{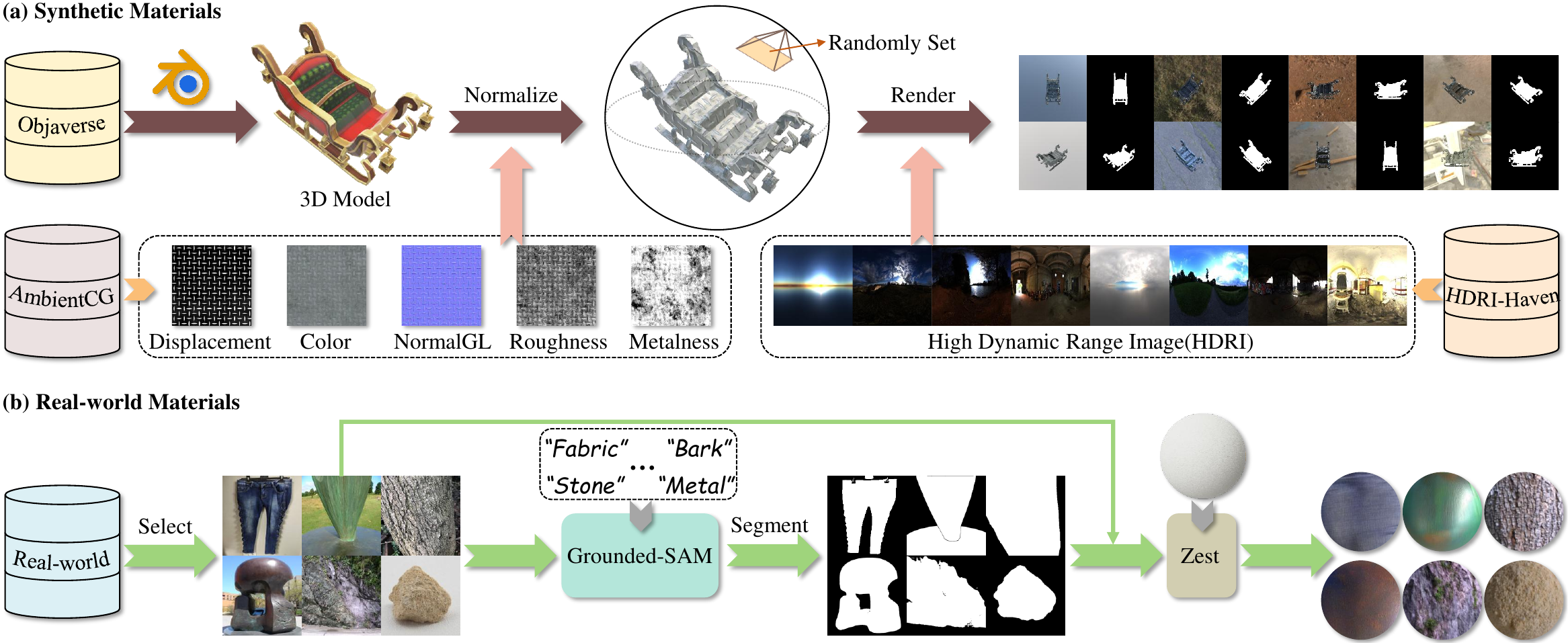}
    \caption{Overview of our dataset construction pipeline. (a) Synthetic materials are generated from 3D models obtained from Objaverse, combined with textures from AmbientCG, and rendered with HDR images. (b) Real-world materials are selected and segmented using Grounded-SAM and then transformed into material spheres via the ZeST method.}
    \label{fig:dataset}
\end{figure*}

\section{Methodology}
In this section, we introduce MaRI, a framework developed to address the domain gap in material retrieval between synthetic and real-world data. To clarify, we distinguish ``image'' as a 2D perspective view and ``material'' as a material ball representation (see Figure\ref{fig:head} (b)), using these terms as abbreviations throughout. MaRI aligns image and material features within a shared feature space \(\mathcal{F}\), allowing for direct comparisons across different domains. To achieve this alignment, we construct a comprehensive dataset \(\mathcal{D} = \{(\mathcal{D}_{\text{synthetic}}, \mathcal{D}_{\text{real}})\}\) with pairs of image and material that combines controlled synthetic samples with diverse real-world materials. Inspired by CLIP, MaRI uses a dual-encoder architecture based on DINOv2, with separate encoders fine-tuned for image and material representations. The focus is on adapting the last Transformer block to retain general visual features while learning domain-specific variations. A contrastive loss guides the training, pulling matched pairs closer in \(\mathcal{F}\) and pushing apart mismatched pairs, making MaRI effective in retrieving materials across both synthetic and real-world settings.

\subsection{Problem Formulation}
Material retrieval involves mapping visual representations and material properties into a shared feature space \(\mathcal{F}\) to enable direct comparison and accurate retrieval. This is achieved through two encoders: \( E_I \) for image space \(\mathcal{X}\) and \( E_M \) for material space \(\mathcal{M}\). Given an image \( x \in \mathcal{X} \) and a material \( m \in \mathcal{M} \), the encoders map these inputs into \(\mathcal{F}\) as \(\mathbf{z}_I = E_I(x)\) and \(\mathbf{z}_M = E_M(m)\), where \( \mathbf{z}_I, \mathbf{z}_M \in \mathcal{F} \) represent the feature embeddings of the image and material, respectively. The mapping facilitates direct comparison of visual features and material attributes within the joint space, making it possible to measure their similarity.

Aligning similar images and materials in the shared space relies on a contrastive learning framework, trained on our constructed dataset \(\mathcal{D} = \{(\mathcal{D}_{\text{synthetic}}, \mathcal{D}_{\text{real}})\}\). The dataset, comprising both synthetic and real-world material samples, provides the model with a diverse range of materials, supporting improved generalization. The objective is to maximize \(\text{sim}(\mathbf{z}_I, \mathbf{z}_M)\) for positive pairs of image embeddings \(\mathbf{z}_I\) and material embeddings \(\mathbf{z}_M\), while minimizing \(\text{sim}(\mathbf{z}_I, \mathbf{z}_{M'})\) for negative pairs \(\mathbf{z}_{M'}\). The shared feature space \(\mathcal{F}\) provides a structure for formulating the material retrieval task as a nearest-neighbor search. For a query image \( x_q \), the objective is to find the material \( m^* \) that maximizes the similarity with the query's feature embedding:

\begin{equation}
m^* = \arg\max_{m \in \mathcal{M}} \text{sim}(\mathbf{z}_{I_q}, \mathbf{z}_M).
\label{eq:nearest_neighbor}
\end{equation}
Through the training of the encoders \( E_I \) and \( E_M \), the representations in \(\mathcal{F}\) are aligned, supporting accurate and efficient material retrieval.

\subsection{Dataset}

Addressing the domain gap between synthetic and real-world materials requires a large-scale dataset that captures a broad range of material types and environmental conditions. To achieve this, we construct a dataset \(\mathcal{D} = \{(\mathcal{D}_{\text{synthetic}}, \mathcal{D}_{\text{real}})\}\) to offer a rich training resource for material retrieval. The dataset is composed of two main components: \(\mathcal{D}_{\text{synthetic}}\) and \(\mathcal{D}_{\text{real}}\), each designed to cover different aspects of material appearance and variability, contributing to a more effective alignment between synthetic control and real-world complexity, as illustrated in Figure~\ref{fig:dataset}.

\subsubsection{Synthetic Materials}
We aim to create a synthetic dataset by rendering various objects associated with diverse materials in different lighting environments in Blender. The process utilizes 3D models \( O_i \) from Objaverse \cite{deitke2022objaverseuniverseannotated3d} and applies a systematic normalization process to ensure consistency across various rendering conditions. Let \(\mathbf{B}_{\min} = (b_x^{\min}, b_y^{\min}, b_z^{\min})\) and \(\mathbf{B}_{\max} = (b_x^{\max}, b_y^{\max}, b_z^{\max})\) denote the minimum and maximum coordinates of the model's axis-aligned bounding box, with \(\mathbf{s} = \mathbf{B}_{\max} - \mathbf{B}_{\min}\) representing its spatial extent. We define the scaling factor \(\alpha\) as \(\alpha = 1 / \max(s_x, s_y, s_z)\), where \( s_x, s_y, s_z \) are the dimensions of \(\mathbf{s}\). To center the model at the origin, we compute the centroid \(\mathbf{c} = (\mathbf{B}_{\max} + \mathbf{B}_{\min}) / 2\). The normalized model \(\mathbf{O}_i'\) is then obtained as:
\begin{equation}
\mathbf{O}_i' = \frac{1}{\max(s_x, s_y, s_z)} \left(O_i - \frac{\mathbf{B}_{\max} + \mathbf{B}_{\min}}{2}\right).
\end{equation}
The transformation scales the model to fit within a unit cube and centers it at the origin, providing uniformity across models and facilitating consistent rendering conditions.

\begin{figure*}[!t]
    \centering
    \includegraphics[width=\textwidth]{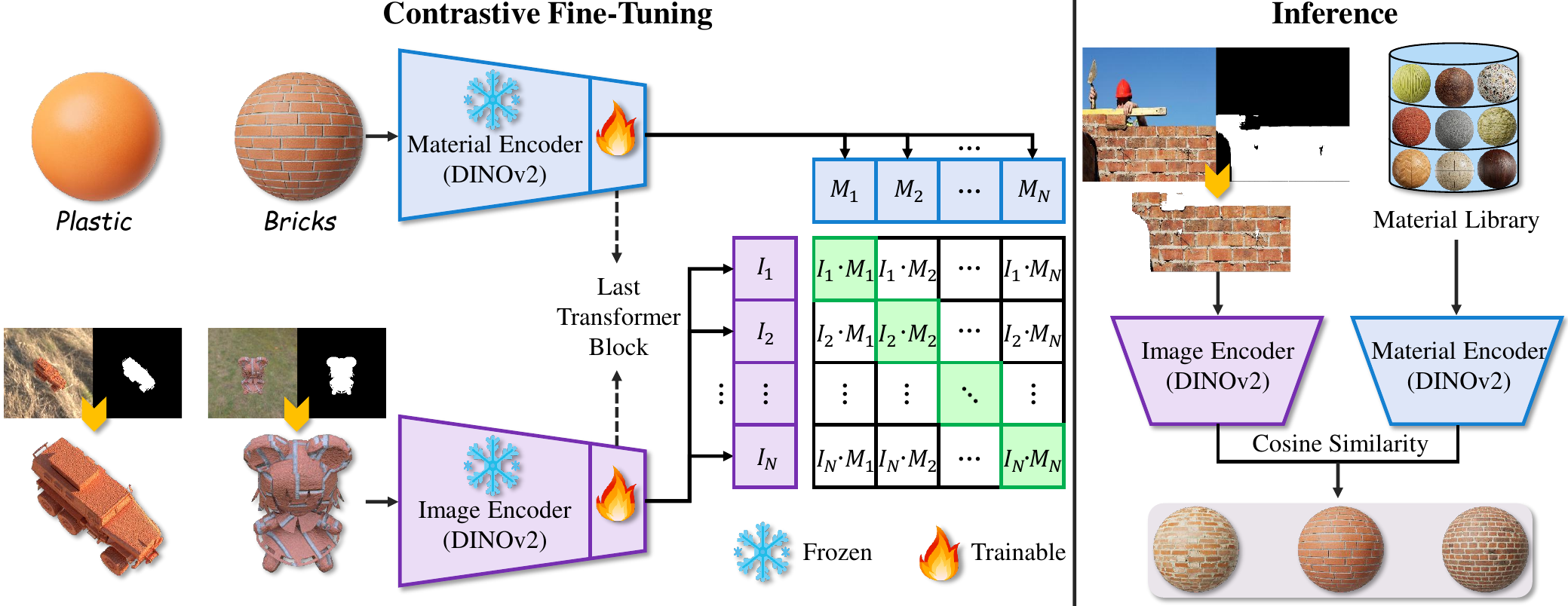}
    \caption{The architecture of the MaRI framework for contrastive fine-tuning in material retrieval. MaRI uses DINOv2-based encoders for both image and material feature extraction, fine-tuning only the last Transformer block, while keeping the rest of the model frozen. During inference, cosine similarity between image and material embeddings is used to retrieve the most relevant materials from the library.}
    \label{fig:framework}
\end{figure*}

Next, we apply materials \( m_j \in \mathcal{M} \) to each model. These materials are drawn from a library of 1605 physically-based rendering (PBR) textures sourced from AmbientCG \cite{ambientcg2023}, covering 86 distinct material categories. The library is designed to encompass most material types. Each material includes Base Color, Normal Map, Roughness, Displacement maps, and Metalness(optional) which are integrated into a principled BSDF shader to simulate realistic surface interactions with light. The materials are represented as tuples of physical properties, capturing diverse physical properties crucial for accurate material representation.

Lighting conditions are varied using 712 HDRI files \( H_k \) sourced from HDRI Haven \cite{hdri_haven}, simulating diverse real-world lighting scenarios. Cameras \( C_l \) are randomly positioned on a hemispherical surface of radius \( r = 3 \) units around each model, with latitude \(\theta\) and longitude \(\phi\) angles sampled as \(\theta \in [5^\circ, 75^\circ]\) and \(\phi \in [-180^\circ, 180^\circ]\). The upper hemisphere placement reduces shadows, while the random positions increase shape variance by capturing multiple perspectives. Each camera's position is calculated by:

\begin{equation}
(x, y, z) = r (\sin \theta \cos \phi, \sin \theta \sin \phi, \cos \theta).
\label{eq:camera_position}
\end{equation}

For each model and material combination, we generate 8 different viewpoints using randomly positioned cameras, with lighting conditions also randomized through different HDRI environments. Each rendering generates an image \( x_i \), its corresponding mask \(\text{mask}_i\), and the applied material descriptor \( m_i \), where the mask delineates the object’s shape within the image. The complete synthetic dataset is defined as: $\mathcal{D}_{\text{synthetic}} = \{(x_i, \text{mask}_i, m_i)\}_{i=1}^{N_{\text{synthetic}}}$, where \( N_{\text{synthetic}} = 394560 \) represents the total number of samples. The dataset covers a wide range of shapes, textures, and lighting conditions, offering a diverse and controlled resource for the material retrieval task. 
% \commentjw{It seems that it is conflicted with the introduction, which states that synthetic data cannot fully capture the richness and variability of real-world appearance.}
\subsubsection{Real-world Materials}

Blender-based synthetic rendering produces high-quality, diverse material samples, yet occasionally encounters a domain gap when applied to real-world material retrieval tasks. Additionally, even though the synthetic data covers a vast majority of material types, the sheer diversity of materials in the real world means that many are still underrepresented. 

Inspired by the ZeST \cite{cheng2024zestzeroshotmaterialtransfer} method's ability to transfer material appearances from real-world images onto neutral material spheres, we expanded our dataset to include a wider variety of real-world materials. We first curated a dataset comprising thousands of real-world images, collected from online sources and various datasets \cite{kaggle_datasets, sculptures6k_2024, Liu-CVPR-10, cvmart}. Priority was then given to images with clearly identifiable foreground objects, which were segmented using the Grounded SAM model \cite{ren2024groundedsamassemblingopenworld} with material-specific prompts to produce accurate object masks. Each image, along with its segmentation mask, was processed through the ZeST pipeline to generate a standardized material representation on a neutral sphere.

The real-world materials dataset also stores three components for each sample: the original image \( x_i \), the segmentation mask \(\text{mask}_i\), and the rendered material sphere \( m_i \). This results in a structured dataset: $\mathcal{D}_{\text{real}} = \{(x_i, \text{mask}_i, m_i)\}_{i=1}^{N_{\text{real}}}$, where \( N_{\text{real}} = 30,000 \). The dataset mainly covers 8 material categories, such as metals, fabrics, woods, and ceramics. By integrating these real-world samples, our dataset can effectively reduce the domain gap between synthetic and real-world materials.

\subsection{Domain-Adaptive Contrastive Learning}
Building on the diverse range of material data in \(\mathcal{D}\), our proposed MaRI framework is inspired by the contrastive learning approach of CLIP, but adapts it to bridge the domain gap between synthetic and real-world visual representations—a key challenge in material retrieval tasks. Rather than aligning different modalities, MaRI focuses on aligning varied visual features within a single modality but across different domains. It utilizes two DINOv2-based encoders, \( E_I \) for masked image representations and \( E_M \) for material properties, to project masked image inputs \( x_i \) and material spheres \( m_i \) into a shared feature space \(\mathcal{F}\):

\begin{equation}
\mathbf{z}_I^i = E_I(x_i \odot \text{mask}_i), \quad \mathbf{z}_M^i = E_M(m_i).
\end{equation}
Here, \(\mathbf{z}_I^i, \mathbf{z}_M^i \in \mathbb{R}^d\) are the embeddings of the masked rendered image and the material sphere image, and \(\odot\) denotes element-wise multiplication to apply the mask. 

We fine-tune only the last Transformer block of each encoder, allowing the model to capture domain-specific variations in materials while retaining the general pre-trained features of DINOv2. As shown in Figure~\ref{fig:framework}, the similarity between the image and material embeddings is computed using a scaled dot-product function:

\begin{equation}
\text{sim}(\mathbf{z}_I^i, \mathbf{z}_M^j) = \frac{\mathbf{z}_I^i \cdot \mathbf{z}_M^j}{\sqrt{d}},
\end{equation}
where \(d\) is the dimensionality of the feature space. We then use the InfoNCE loss with a temperature parameter $\tau=0.07$, which controls the sharpness of the resulting distribution, to pull the representations of matching pairs closer while pushing non-matching pairs apart in the shared feature space:

\begin{equation}
\mathcal{L}_{\text{contrast}} = - \frac{1}{N} \sum_{i=1}^{N} \log 
  \frac{\exp\bigl(\text{sim}(\mathbf{z}_I^i, \mathbf{z}_M^i)/\tau\bigr)}
       {\sum_{j=1}^{N} \exp\bigl(\text{sim}(\mathbf{z}_I^i, \mathbf{z}_M^j)/\tau\bigr)}.
\end{equation}
The loss encourages positive pairs \((\mathbf{z}_I^i, \mathbf{z}_M^i)\) to have higher similarity scores than any other pair \((\mathbf{z}_I^i, \mathbf{z}_M^j)\) with \( j \neq i \), thus aligning the features in a domain-agnostic manner. MaRI effectively creates a shared space that supports robust material retrieval across varying data sources.

\section{Experiments}
In this section, we conduct a comprehensive evaluation of the MaRI framework for material retrieval tasks. Section~\ref{sec:datasets_metrics} introduces the test datasets and evaluation metrics employed in our experiments, providing context for the subsequent analyses. Section~\ref{sec:comparison} presents a comparative analysis, where MaRI's performance is benchmarked against existing models commonly used for material search. We also explore, in Section~\ref{sec:ablation}, how key design elements like model architecture, training strategies, and data composition influence MaRI's results. Finally, Section~\ref{qualitative results} provides additional qualitative results, demonstrating MaRI’s capacity to retrieve accurate matches from the Unseen Materials dataset.

\subsection{Test Datasets and Metrics}
\label{sec:datasets_metrics}

To evaluate the effectiveness of MaRI, we design two distinct test datasets and corresponding evaluation protocols to assess material retrieval performance. 

The first test dataset, marked as ``Trained'' in all experiments, evaluates material retrieval performance using novel test images from a material gallery derived from AmbientCG. Specifically, we selected approximately 200 materials from the synthetic dataset, spanning mainly eight categories: wood, metal, plastic, leather, fabric, stone, ceramic, and rubber. For each selected material, corresponding real-world images were collected online, and material-specific regions were annotated with segmentation masks. Retrieval was performed using these real-world images as queries, with the target being the original 200 materials in the gallery of 1605 synthetic materials. We report the following metrics: (1) Top-1 and Top-5 instance-level accuracy(T1I and T5I), which measure the ability of the model to precisely retrieve the correct material from the gallery; (2) Top-1 class-level accuracy(T1C), assessing how accurately the model classifies materials into predefined categories; and  
(3) Top-3 Intersection-over-Union (T3IoU), which quantifies the overlap between the predicted material categories and the ground truth, providing a measure of category-level alignment. The use of Top-1 and Top-3 metrics is driven by the inherent scarcity of certain materials in the gallery, as higher Top-k metrics may include irrelevant matches due to the limited number of materials closely resembling the target.

The second test dataset is marked as ``Unseen'' in all experiments, which evaluates generalization to novel test materials by utilizing a new gallery of approximately 200 materials curated from the Textures website \cite{textures_2024}, which were unseen during the training process. Similar to the previous setup, for each material, real-world images and their corresponding material segmentation masks were collected, and retrieval was conducted within this newly constructed gallery. This scenario evaluates MaRI's ability to retrieve accurate matches for real-world queries from a gallery of previously unseen materials. Due to the diverse and unstructured nature of the material categories in this dataset, the evaluation emphasizes Top-1 and Top-5 instance-level accuracy, showcasing the framework's capability to identify the most relevant matches.

\begin{figure*}[t]
\centering
\includegraphics[width=\linewidth]{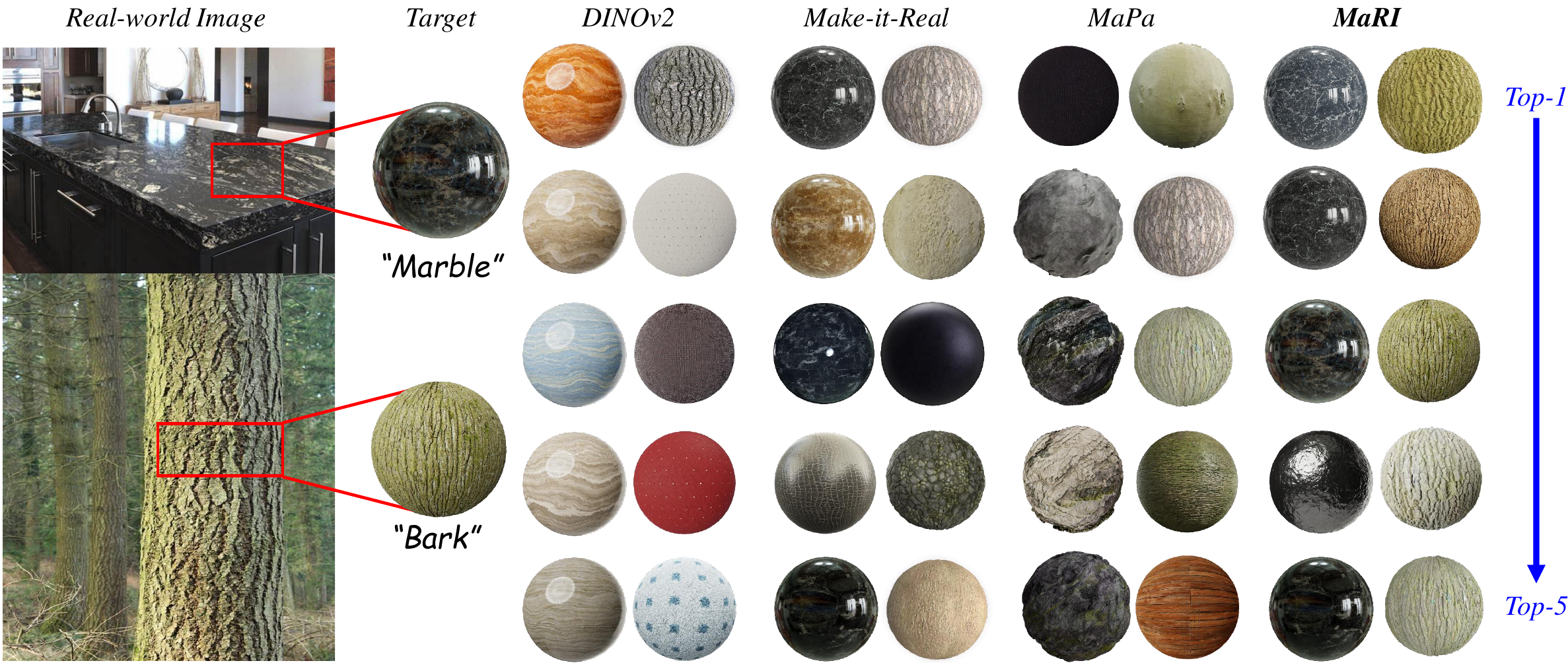}
\caption{Qualitative comparison of material retrieval results using the Trained Materials dataset as the gallery.}
\label{fig:qualitative_comparison}
\vspace{-10pt}
\end{figure*}

\subsection{Comparative Analysis of Material Retrieval}
\label{sec:comparison}
Given the novelty of the material retrieval task, there are currently no directly comparable methods designed specifically for this purpose. Our comparisons therefore draw on related approaches, including general-purpose image search models like ViT, CLIP, and DINOv2, which serve as baselines for instance- and class-level retrieval evaluations. Additionally, we compare MaRI against existing material retrieval approaches used in other works. Make-it-Real leverages GPT-4V to perform hierarchical material searches, starting with high-level category identification followed by detailed matching within a structured material library. MaPa integrates GPT-4V with CLIP by first performing coarse material categorization using GPT-4V and then refining the search within the predicted category using CLIP for detailed material retrieval. 
\begin{table}[htbp]
\caption{Material Retrieval Performance on Trained and Unseen Datasets. Best values are highlighted in \textcolor{lakeblue}{\textbf{blue}}.}
\centering
\resizebox{\linewidth}{!}{%
\begin{tabular}{c|cccc|cc}
\toprule
\multirow{2}{*}{\textbf{Method}} & \multicolumn{4}{c|}{\textbf{Trained}} & \multicolumn{2}{c}{\textbf{Unseen}} \\
\cmidrule(lr){2-5} \cmidrule(lr){6-7}
 & T1I & T5I & T1C & T3IoU & T1I & T5I \\
\midrule
ViT \cite{dosovitskiy2021imageworth16x16words}          & 3.5\%  & 12.0\% & 16.0\% & 0.41 & 16.5\% & 56.0\% \\
DINOv2 \cite{oquab2024dinov2learningrobustvisual}      & 7.5\%  & 28.0\% & 69.0\% & 0.67 & 31.0\% & 62.5\% \\
CLIP \cite{radford2021learningtransferablevisualmodels}         & 2.0\%  & 11.0\% & 36.5\% & 0.47 & 14.0\% & 29.5\% \\
Make-it-Real \cite{fang2024makeitrealunleashinglargemultimodal} & 8.5\%  & 16.0\% & 76.5\% & 0.60 & 42.5\% & 75.0\% \\
MaPa \cite{zhang2024mapatextdrivenphotorealisticmaterial}         & 2.5\%  & 17.5\% & 80.0\% & \textcolor{lakeblue}{\textbf{0.80}} & 19.5\% & 69.0\% \\
MaRI         & \textcolor{lakeblue}{\textbf{26.0\%}} & \textcolor{lakeblue}{\textbf{90.0\%}} & \textcolor{lakeblue}{\textbf{81.5\%}} & 0.77 & \textcolor{lakeblue}{\textbf{54.0\%}} & \textcolor{lakeblue}{\textbf{89.0\%}} \\
\bottomrule
\end{tabular}
}
\label{tab:comparison}
\vspace{-8pt}
\end{table}

Table~\ref{tab:comparison} highlights the material retrieval performance across both known and unseen galleries. In the Trained Materials, MaRI achieves Top-1 instance accuracy of 26.0\%, Top-5 instance accuracy of 90.0\%, and Top-1 class accuracy of 81.5\%, outperforming all other methods. Although MaRI's Top-3 IoU score of 0.77 is slightly lower than MaPa's 0.80, this difference arises from the distinct retrieval processes employed by the two methods. MaPa utilizes GPT-4V for coarse-grained classification into material categories before conducting a fine-grained search within the same category. As a result, its IoU scores remain consistent across Top-k predictions and directly align with its Top-1 class accuracy. In contrast, MaRI performs retrieval directly within a unified embedding space, enabling the discovery of visually similar materials that may belong to different categories. Although the expanded search scope can occasionally result in category mismatches among the Top-k results, causing a minor decrease in Intersection over Union (IoU) compared to MaPa, it offers substantial benefits when it comes to retrieving materials at the instance level, outperforming all other methods in this regard. In the unseen materials gallery, MaRI also achieves significant improvements, with Top-1 and Top-5 instance accuracies of 54.0\% and 89.0\%, respectively, far surpassing Make-it-Real (42.5\% and 75.0\%). These results showcase MaRI's superior performance across both known and unseen material retrieval tasks.

The qualitative results in Figure~\ref{fig:qualitative_comparison} illustrate that general image search models, such as the original DINOv2, struggle to capture the intricate relationships between material textures and their corresponding images. As previously demonstrated through quantitative evaluations, methods like CLIP and ViT exhibit similarly poor performance and are therefore omitted from this figure. In contrast, MaRI also outperforms GPT-4V-based material retrieval methods, including MaPa and Make-it-Real, by more effectively capturing fine-grained material characteristics. For instance, in the ``Bark'' case shown in Figure~\ref{fig:qualitative_comparison}, MaRI consistently retrieves materials in its Top-5 predictions that closely resemble the target in both texture and color.

\subsection{Ablation Studies}
\label{sec:ablation}
\textbf{Impact of Synthetic Data Scale.}
The complexity of texture and material information necessitates a large dataset for contrastive learning to capture material features effectively and enable accurate retrieval within the embedding space. We conducted an ablation study to evaluate the impact of synthetic data scale on MaRI's performance, as detailed in Table~\ref{tab:ablation_data_usage}. The findings in Table~\ref{tab:ablation_data_usage} illustrate a strong correlation between the scale of synthetic data and the improvement in instance-level retrieval accuracy.
\begin{table}[t]
\caption{Ablation study evaluating the impact of synthetic data usage. Best values are highlighted in \textcolor{lakeblue}{\textbf{blue}}.}
\centering
\small
\resizebox{\linewidth}{!}{%
\begin{tabular}{c|cccc|cc}
\toprule
\multirow{2}{*}{\textbf{Data \%}} & \multicolumn{4}{c|}{\textbf{Trained}} & \multicolumn{2}{c}{\textbf{Unseen}} \\
\cmidrule(lr){2-5} \cmidrule(lr){6-7}
 & T1I & T5I & T1C & T3IoU & T1I & T5I \\
\midrule
25\%   & 19.5\% & 55.5\% & 77.5\% & 0.76 & 44.5\% & 83.5\% \\
50\%   & 20.0\% & 63.5\% & \textcolor{lakeblue}{\textbf{82.0\%}} & \textcolor{lakeblue}{\textbf{0.79}} & 46.0\% & 85.5\% \\
75\%   & 22.0\% & 79.5\% & 80.5\% & 0.78 & 48.5\% & 80.0\% \\
100\%  & \textcolor{lakeblue}{\textbf{26.0\%}} & \textcolor{lakeblue}{\textbf{90.0\%}} & 81.5\% & 0.77 & \textcolor{lakeblue}{\textbf{54.0\%}} & \textcolor{lakeblue}{\textbf{89.0\%}} \\
\bottomrule
\end{tabular}
}
\label{tab:ablation_data_usage}
\vspace{-15pt}
\end{table}
For the Trained Materials dataset, Top-1 instance accuracy increases from 19.5\% with 25\% of the data to 26.0\% with the full dataset, and Top-5 instance accuracy sees a substantial rise from 55.5\% to 90.0\%. Similarly, in the Unseen Materials dataset, Top-1 instance accuracy improves from 44.5\% to 54.0\%, showcasing MaRI's enhanced capability to generalize to previously unseen materials. Interestingly, while instance-level retrieval improves consistently with data scale, the Top-1 class-level accuracy exhibits relatively smaller gains, peaking at 82.0\% for 50\% of the dataset and slightly decreasing to 81.5\% with the full dataset. The plateau suggests that class-level classification may benefit less from additional synthetic data due to the saturation of categorical information in the dataset. The observations highlight how a larger and more diverse dataset contributes to overall performance improvements, especially in instance-level retrieval.\\
\textbf{Model Architecture and Data Composition.}
Building on the findings regarding the significance of synthetic data scale, we further analyze the contributions of key architectural components and data composition in optimizing MaRI's performance. As demonstrated in Table~\ref{tab:ablation_accuracy}, the combination of dual encoders with both synthetic and real-world data achieves the highest retrieval accuracies. The results validate the theoretical premise that the material and image spaces represent distinct domains, and employing dual encoders effectively reduces the domain gap by learning separate representations for each space while aligning them in the shared embedding space. Training with both synthetic and real data outperforms using either dataset alone. For instance, in the Trained Materials dataset, the combination of synthetic and real data achieves the highest Top-1 instance accuracy of 26.0\%. Removing synthetic or real data significantly reduces instance-level accuracy, with Top-5 instance accuracy dropping from 90.0\% to 62.0\% or 27.5\%, respectively. Additionally, the Unseen Materials dataset further underscores the importance of real data in enhancing generalization. Training with both datasets yields a Top-1 instance accuracy of 54.0\%, compared to 44.0\% or 35.0\% when excluding synthetic or real data. Leveraging both datasets allows MaRI to establish stronger relationships between material and image spaces, resulting in robust performance across diverse and previously unseen materials.\\
\textbf{Fine-Tuning Configurations.}
An analysis of Table~\ref{tab:ablation_loss} reveals that fine-tuning only the final Transformer block of DINOv2, while freezing other parameters, consistently yields better results across both InfoNCE and Triplet loss configurations. This is because the early layers of the pre-trained DINOv2 model capture generalizable low-level and mid-level features critical for material and image representations. Freezing these layers prevents overfitting to the training dataset and retains the model's ability to generalize across diverse material distributions. Under the same configurations, InfoNCE loss demonstrates superior performance over Triplet loss due to its ability to optimize a batch-wise similarity matrix, which evaluates all material-image pairs simultaneously. It allows the model to capture more nuanced relationships within the embedding space, effectively aligning material and image features. In contrast, Triplet loss focuses on individual anchor-positive-negative triplets, which limits its capacity to fully explore the complex associations.
\begin{table}[t]
\caption{Ablation study on model architecture and data composition. \cmark\ indicates the component is enabled, while \xmark\ indicates it is disabled. Best values are highlighted in \textcolor{lakeblue}{\textbf{blue}}. Abbreviations: DE = Dual Encoder, RD = Real Data, SD = Synthetic Data.}
\centering
\small 
\resizebox{\linewidth}{!}{%
\begin{tabular}{ccc|cccc|cc}
\toprule
\multicolumn{3}{c|}{\textbf{Configuration}} & \multicolumn{4}{c|}{\textbf{Trained}} & \multicolumn{2}{c}{\textbf{Unseen}} \\
\cmidrule(lr){1-3} \cmidrule(lr){4-7} \cmidrule(lr){8-9}
\textbf{DE} & \textbf{RD} & \textbf{SD} & \textbf{T1I} & \textbf{T5I} & \textbf{T1C} & \textbf{T3IoU} & \textbf{T1I} & \textbf{T5I} \\
\midrule
\cellcolor{lakegreen}\cmark & \xmark & \cellcolor{lakegreen}\cmark & 20.5\% & 62.0\% & 75.5\% & 0.76 & 44.0\% & 78.0\% \\
\cellcolor{lakegreen}\cmark & \cellcolor{lakegreen}\cmark & \xmark & 9.0\% & 27.5\% & 45.0\% & 0.49 & 35.0\% & 63.5\% \\
\xmark & \cellcolor{lakegreen}\cmark & \cellcolor{lakegreen}\cmark & 20.5\% & 61.0\% & 77.5\% & 0.74 & 49.5\% & 85.5\% \\
\cellcolor{lakegreen}\cmark & \cellcolor{lakegreen}\cmark & \cellcolor{lakegreen}\cmark & \textcolor{lakeblue}{\textbf{26.0\%}} & \textcolor{lakeblue}{\textbf{90.0\%}} & \textcolor{lakeblue}{\textbf{81.5\%}} & \textcolor{lakeblue}{\textbf{0.77}} & \textcolor{lakeblue}{\textbf{54.0\%}} & \textcolor{lakeblue}{\textbf{89.0\%}} \\
\bottomrule
\end{tabular}
}
\label{tab:ablation_accuracy}
\vspace{-13pt}
\end{table}

\begin{table}[t]
\caption{Ablation study on fine-tuning configurations. \cmark\ indicates the component is enabled, while \xmark\ indicates it is disabled. Best values are highlighted in \textcolor{lakeblue}{\textbf{blue}}. Abbreviations: LBO = Last Block Only, TL = Triplet Loss, IL = InfoNCE Loss.}
\centering
\small
\resizebox{\linewidth}{!}{%
\begin{tabular}{ccc|cccc|cc}
\toprule
\multicolumn{3}{c|}{\textbf{Configuration}} & \multicolumn{4}{c|}{\textbf{Trained}} & \multicolumn{2}{c}{\textbf{Unseen}} \\
\cmidrule(lr){1-3} \cmidrule(lr){4-7} \cmidrule(lr){8-9}
\textbf{LBO} & \textbf{TL} & \textbf{IL} & \textbf{T1I} & \textbf{T5I} & \textbf{T1C} & \textbf{T3IoU} & \textbf{T1I} & \textbf{T5I} \\
\midrule
\xmark & \xmark & \cellcolor{lakegreen}\cmark & 13.0\% & 42.5\% & 69.0\% & 0.65 & 31.5\% & 67.0\% \\
\xmark & \cellcolor{lakegreen}\cmark & \xmark & 7.5\% & 21.0\% & 53.0\% & 0.49 & 15.5\% & 52.5\% \\
\cellcolor{lakegreen}\cmark & \cellcolor{lakegreen}\cmark & \xmark & 5.5\% & 31.5\% & 73.0\% & 0.71 & 38.5\% & 71.5\% \\
\cellcolor{lakegreen}\cmark & \xmark & \cellcolor{lakegreen}\cmark & \textcolor{lakeblue}{\textbf{26.0\%}} & \textcolor{lakeblue}{\textbf{90.0\%}} & \textcolor{lakeblue}{\textbf{81.5\%}} & \textcolor{lakeblue}{\textbf{0.77}} & \textcolor{lakeblue}{\textbf{54.0\%}} & \textcolor{lakeblue}{\textbf{89.0\%}} \\
\bottomrule
\end{tabular}
}
\label{tab:ablation_loss}
\end{table}

\subsection{More Qualitative Results}
\label{qualitative results}
\begin{figure}[t]
    \centering
    \includegraphics[width=\linewidth]{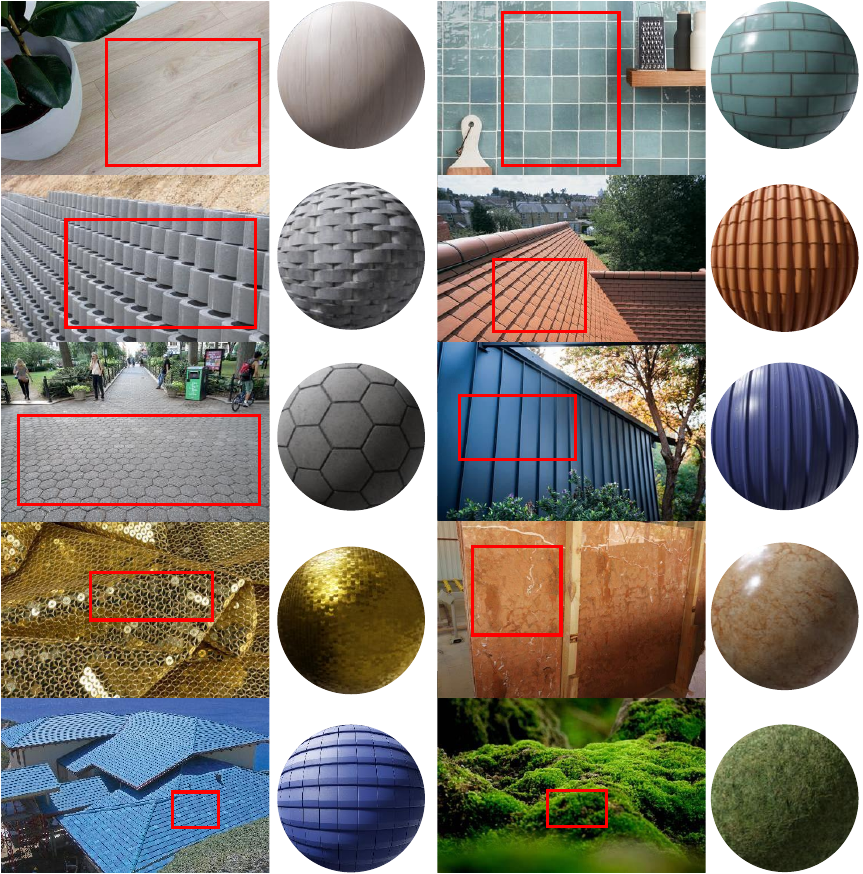}
    \caption{Examples of Top-1 material retrieval results using the Unseen Materials gallery as the search space.}
    \label{fig:unseen_materials}
    \vspace{-15pt}
\end{figure}

The retrieval results in Figure~\ref{fig:unseen_materials} highlight MaRI's effectiveness in identifying visually similar materials from the Unseen Materials dataset. On the left side, each example presents a real-world image, with the corresponding material sphere retrieved from the unseen gallery shown on the right. MaRI successfully captures fine-grained details, such as the textured surface of tiles, the ruggedness of brick patterns, and the organic structure of moss. These examples demonstrate MaRI’s strong generalization across diverse material.

\section{Conclusion}
We introduce MaRI, a novel framework designed specifically to address the challenges of material retrieval by aligning image and material features in a shared embedding space. A key component of MaRI is the construction of a comprehensive dataset, integrating both synthetic and real-world materials to effectively bridge the domain gap. MaRI successfully captures essential material properties and achieves strong generalization to unseen materials. Unlike prior methods, MaRI provides an effective framework for tackling the unique challenges of material retrieval, achieving strong performance in diverse scenarios. 
{
    \small
    \bibliographystyle{ieeenat_fullname}
    \bibliography{main}
}

\clearpage
\setcounter{section}{0}
\setcounter{table}{0}
\setcounter{figure}{0}
\maketitlesupplementary

This supplementary material provides additional details and insights into the MaRI framework. Section~\ref{sec:implementation} outlines the implementation details, including training configurations and the use of pre-trained DINOv2 \cite{oquab2024dinov2learningrobustvisual} backbones. Section~\ref{sec:backbone_variations} presents an analysis of how different backbone architectures impact material retrieval performance on both trained and unseen datasets. Practical applications of MaRI, such as assigning materials to 3D models for design workflows, are demonstrated in Section~\ref{sec:applications}, showcasing the system’s versatility and ease of use. In addition, Section~\ref{sec:user_study} reports a user study, and Section~\ref{sec:discussion} offers further discussion on limitations, and potential future improvements.

\section{Implementation Details}
\label{sec:implementation} 
We use DINOv2 as the backbone for both the image and material encoders, initialized with pre-trained weights. The training process consists of two main stages: first, the model is fine-tuned on the synthetic dataset for 1 epoch using the Adam optimizer with a learning rate of \(1 \times 10^{-4}\). This initial phase helps establish a robust baseline. Next, the model undergoes fine-tuning on the real-world dataset for 25 epochs at a reduced learning rate of \(1 \times 10^{-5}\) to enhance generalization by capturing more intricate real-world features. To facilitate effective alignment in the shared feature space, the temperature parameter \(\tau\) in the contrastive loss is set to 0.07. The dataset is divided into 90\% for training and 10\% for validation, and a batch size of 256 is employed. The entire process is conducted on four NVIDIA A100 GPUs (80GB each), completing within 3 hours.

\section{Effect of Backbone Variations}
\label{sec:backbone_variations}
The performance of different backbone architectures for material retrieval is summarized in Table~\ref{tab:backbone_comparison}. DINOv2 demonstrates the strongest overall performance, achieving top-1 instance accuracy (T1I) of 26.0\% and top-5 instance accuracy (T5I) of 90.0\% on the Trained dataset, along with 54.0\% and 89.0\% on the Unseen dataset, respectively.

\begin{table}[htbp]
\caption{Backbone comparison for material retrieval on Trained and Unseen datasets. Best values are highlighted in \textcolor{lakeblue}{\textbf{blue}}.}
\centering
\resizebox{\linewidth}{!}{%
\begin{tabular}{c|cccc|cc}
\toprule
\multirow{2}{*}{\textbf{Backbone}} & \multicolumn{4}{c|}{\textbf{Trained}} & \multicolumn{2}{c}{\textbf{Unseen}} \\
\cmidrule(lr){2-5} \cmidrule(lr){6-7}
 & T1I & T5I & T1C & T3IoU & T1I & T5I \\
\midrule
ResNet50 \cite{he2015deepresiduallearningimage}              & 7.5\% & 28.0\% & 64.5\% & 0.58 & 30.0\% & 60.5\% \\
ViT \cite{dosovitskiy2021imageworth16x16words}     & 15.0\% & 33.5\% & 54.5\% & 0.67 & 21.0\% & 65.0\% \\
Swin Transformer \cite{liu2021swintransformerhierarchicalvision}                    & 14.0\% & 41.5\% & 77.0\% & 0.68 & 38.5\% & 77.0\% \\
ConvNeXt \cite{liu2022convnet2020s}                       & 16.5\% & 42.0\% & 70.0\% & 0.67 & 35.0\% & 73.5\% \\
EfficientNet \cite{tan2020efficientnetrethinkingmodelscaling}              & 8.5\% & 21.5\% & 52.0\% & 0.55 & 22.0\% & 54.0\% \\
CLIP \cite{radford2021learningtransferablevisualmodels}              & 12.0\% & 44.5\% & 61.0\% & 0.68 & 29.5\% & 72.5\% \\
DINOv2 \cite{oquab2024dinov2learningrobustvisual} & \textcolor{lakeblue}{\textbf{26.0\%}} & \textcolor{lakeblue}{\textbf{90.0\%}} & \textcolor{lakeblue}{\textbf{81.5\%}} & \textcolor{lakeblue}{\textbf{0.77}} & \textcolor{lakeblue}{\textbf{54.0\%}} & \textcolor{lakeblue}{\textbf{89.0\%}} \\
\bottomrule
\end{tabular}
}
\label{tab:backbone_comparison}
\vspace{-10pt}
\end{table}

\noindent Table~\ref{tab:backbone_comparison} highlights DINOv2's capability to effectively capture complex material characteristics and generalize across diverse data distributions. Other architectures show moderate performance, reflecting limited generalization capacity. These findings emphasize the critical role of backbone selection in enhancing material retrieval tasks, with DINOv2 emerging as the most effective backbone for bridging the synthetic-to-real domain gap.

\section{Applications}
\label{sec:applications}

Having established the effectiveness of MaRI in retrieving materials accurately, we now highlight a practical application that leverage MaRI’s capabilities in real-world 3D design workflows. MaRI empowers users to effortlessly assign desired materials to different parts of a 3D model by providing a streamlined paradigm for material retrieval and application. For instance, a user working with a 3D chair model can specify preferred materials—such as leather for the seat, metal for the legs, and wood for the armrests—through simple input reference images. MaRI efficiently retrieves matching physically based rendering (PBR) materials and applies them to the respective parts. The workflow involves the following steps:

\begin{enumerate}
    \item \textbf{Object Segmentation:} The 3D model is segmented into distinct components (e.g., seat, legs, and armrests) based on its structural design or user annotations.
    \item \textbf{Material Retrieval:} Users provide reference images for their desired material patterns, such as leather textures or wood grains. MaRI leverages its robust library to retrieve accurate and visually consistent matches.
    \item \textbf{Material Application:} The retrieved materials are mapped and applied to the segmented parts of the model, enabling users to visualize their design with photorealistic fidelity.
\end{enumerate}

\begin{figure*}[t]
\centering
\includegraphics[width=\textwidth]{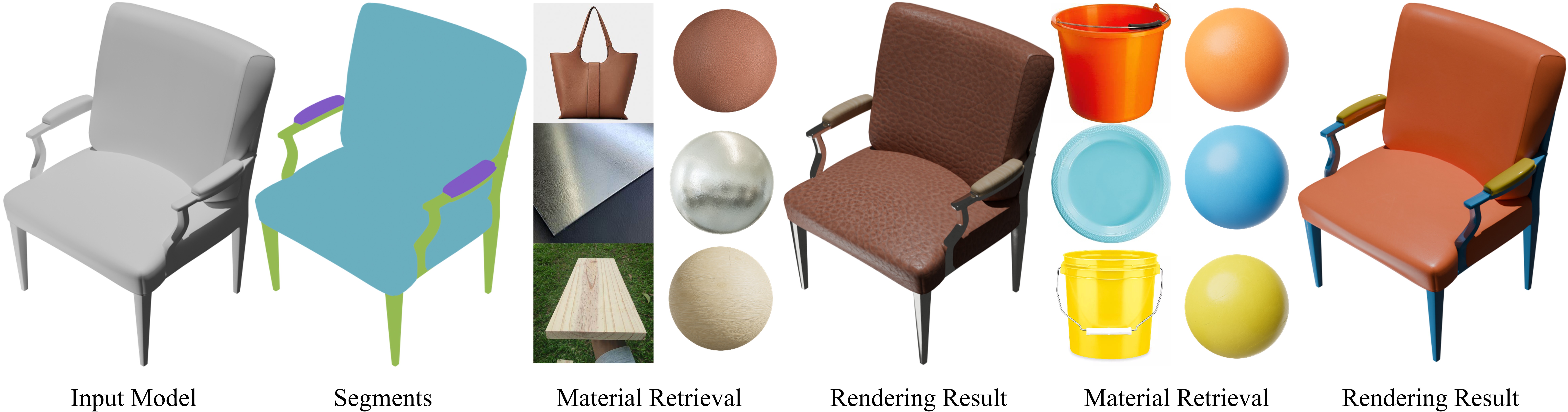}
\caption{Material assignment and rendering for a 3D chair model.}
\label{fig:case1}
\vspace{-10pt}
\end{figure*}

\noindent This process, exemplified through the chair model in Figure~\ref{fig:case1}, highlights MaRI's ability to offer users an intuitive and efficient workflow for realizing their creative goals. MaRI transforms the traditionally labor-intensive process, allowing users to quickly locate and apply desired materials, resulting in outcomes that closely match their creative intent.

\begin{figure*}[htbp]
    \centering
    \includegraphics[width=0.8\textwidth]{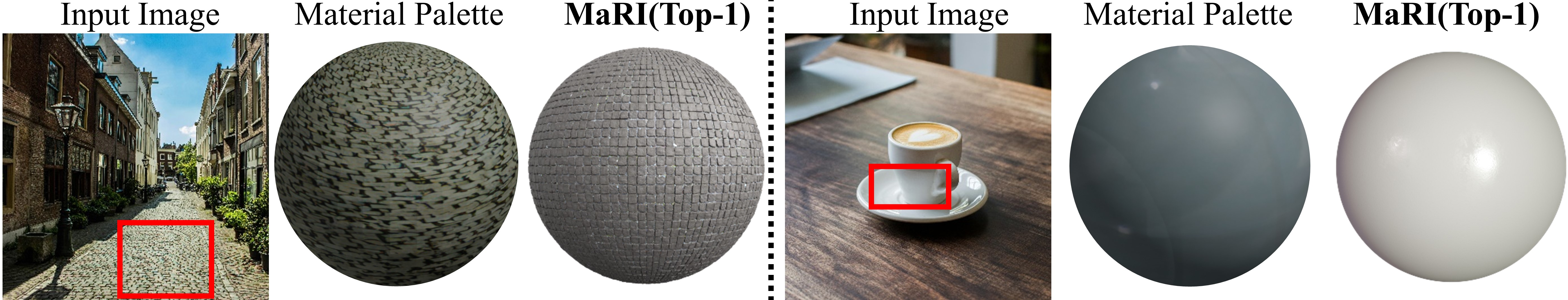}
    \caption{Comparison between Material Palette and MaRI.}
    \label{fig:material_comparison}
    \vspace{-10pt}
\end{figure*}

\section{User Study}
\label{sec:user_study}
We conducted a user study with 20 participants (40\% designers, 30\% researchers, and 30\% general users). Each participant provided three material images as queries, and the top-5 retrievals from MaRI and baseline methods were rated on relevance, realism, and perceptual consistency using a 5-point Likert scale. Table~\ref{tab:user_study} shows that MaRI achieved consistently higher scores across all top-k retrieval results. These results indicate that our method retrieves materials that are more relevant, realistic, and perceptually consistent compared to the baselines.

\begin{table}[t]
\centering
\caption{User study average scores for top-k retrievals. Best values are highlighted in \textcolor{lakeblue}{\textbf{blue}}.}
\begin{tabular}{cccc}
\toprule
\textbf{Method} & \textbf{Top-1} & \textbf{Top-3} & \textbf{Top-5} \\
\midrule
MaPa \cite{zhang2024mapatextdrivenphotorealisticmaterial}             & 2.60 & 3.05 & 3.80 \\
Make-it-Real \cite{fang2024makeitrealunleashinglargemultimodal}     & 3.25 & 3.50 & 3.90 \\
MaRI     & \textcolor{lakeblue}{\textbf{4.15}} & \textcolor{lakeblue}{\textbf{4.50}} & \textcolor{lakeblue}{\textbf{4.85}} \\
\bottomrule
\end{tabular}
\label{tab:user_study}
\vspace{-15pt}
\end{table}

\section{Discussion} 
\label{sec:discussion}
\begin{figure*}[htbp]
\centering
\includegraphics[width=\textwidth]{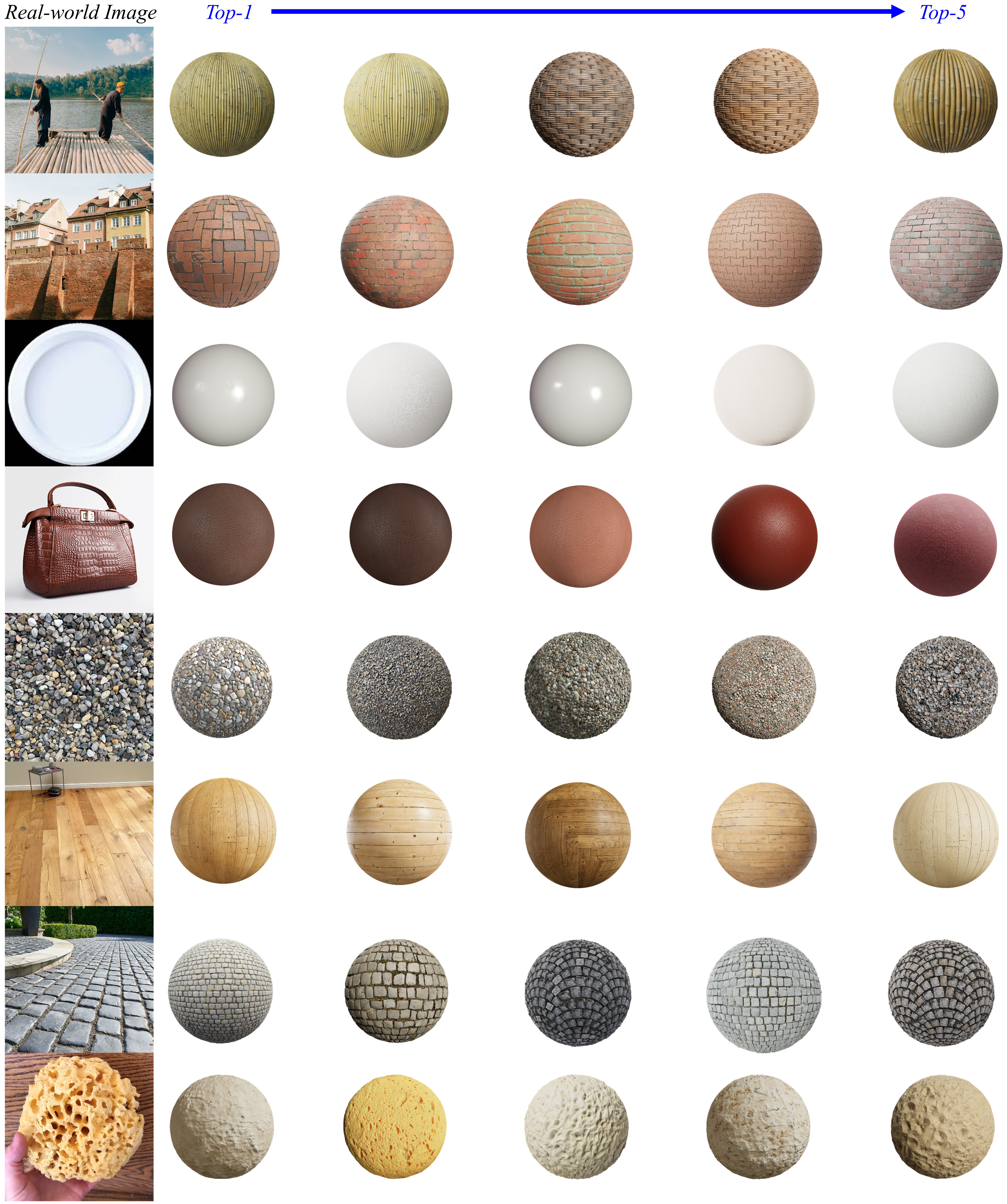}
\caption{Top-5 material retrieval results for real-world images.}
\label{fig:moreresults2}
\end{figure*}

We further analyze the behavior of our retrieval system. As shown in Figure~\ref{fig:moreresults2}, the last row presents a case where a query for sponge material returns several sand-like materials. This outcome is due to the limited number of sponge samples available in our gallery, which causes the system to favor visually similar textures. The case highlights a potential limitation of our approach compared to direct generative methods like Material Palette \cite{lopes2023materialpaletteextractionmaterials}, which can synthesize a broader variety of material appearances. However, our retrieval-based approach is faster than diffusion-based methods. While Material Palette generates novel material images using diffusion, our method retrieves assets from an existing database. This distinction is crucial, as it ensures that our results are not only of high quality but also readily support PBR rendering. Figure~\ref{fig:material_comparison} shows a comparison case.

In our work, we adopt material spheres as the shape representation, which is a widely used choice among content creators. MaRI effectively captures essential material properties and supports robust retrieval performance. We believe that exploring alternative representations, such as blob \cite{10.1145/1275808.1276473} or Havran's \cite{10.5555/3071773.3071775} shapes, may further enhance perceptual alignment \cite{10.1145/3450626.3459813} by capturing subtle material nuances. This presents a promising direction for future research.

\begin{figure*}[!t]
\centering
\includegraphics[width=\textwidth]{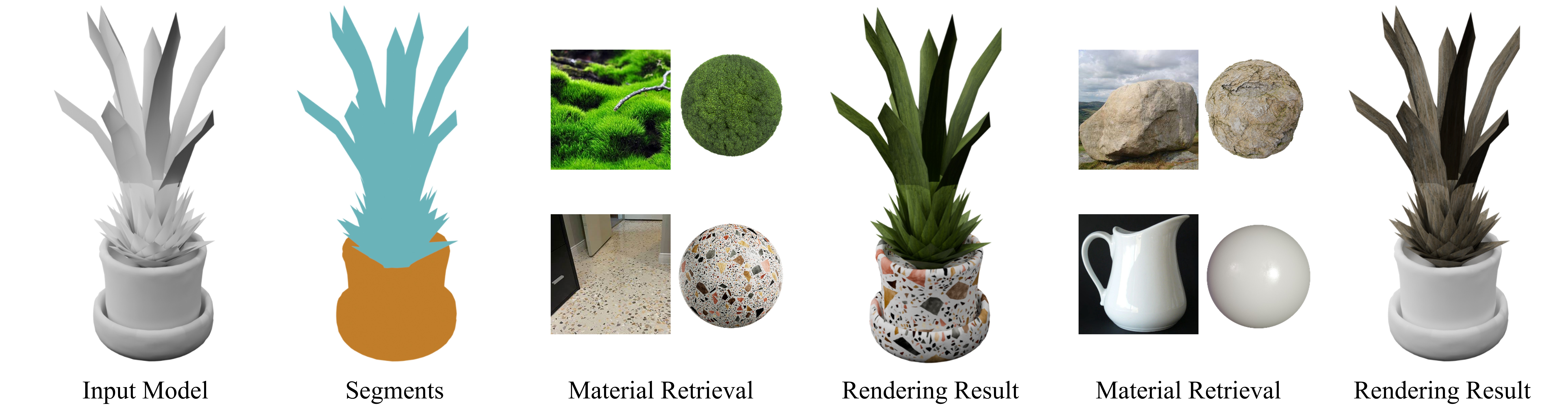}
\caption{Material assignment and rendering for a 3D plant model.}
\label{fig:case2}
\vspace{-10pt}
\end{figure*}

\begin{figure*}[!t]
\centering
\includegraphics[width=\textwidth]{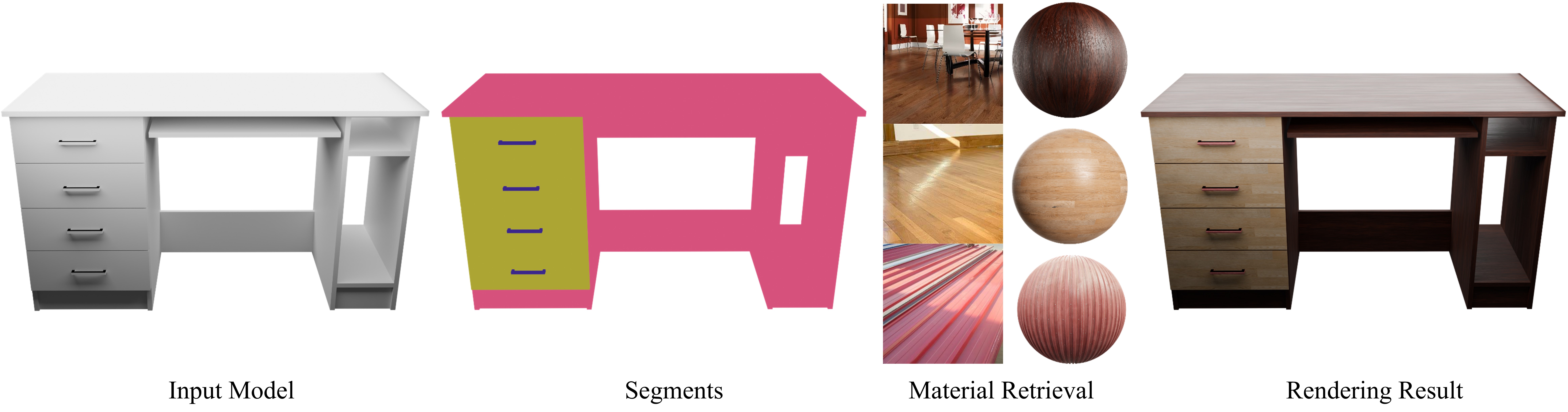}
\caption{Material assignment and rendering for a 3D desk model.}
\label{fig:case3}
\vspace{-10pt}
\end{figure*}

\begin{figure*}[!t]
\centering
\includegraphics[width=\textwidth]{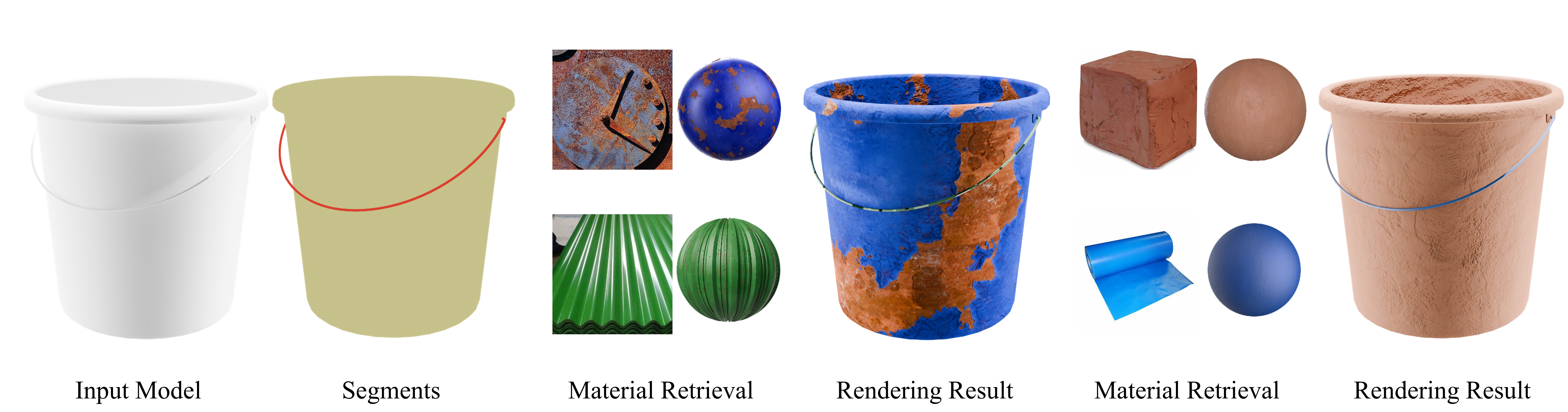}
\caption{Material assignment and rendering for a 3D bucket model.}
\label{fig:case4}
\end{figure*}

\begin{figure*}[htbp]
\centering
\includegraphics[width=\textwidth]{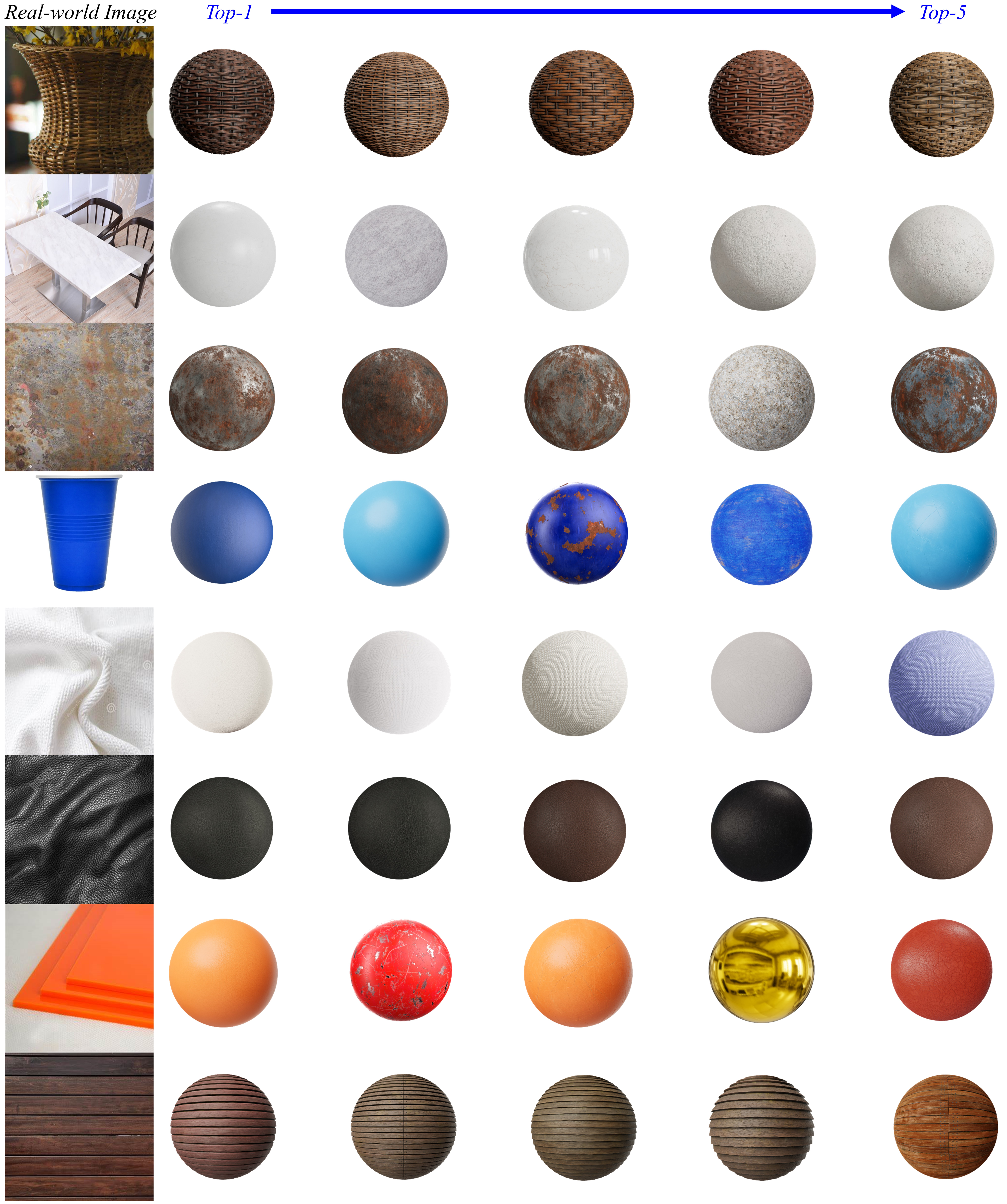}
\caption{Top-5 material retrieval results for real-world images.}
\label{fig:moreresults}
\end{figure*}

\end{document}